\documentclass[10pt,twocolumn,letterpaper]{article}

\usepackage{iccv}
\usepackage{times}
\usepackage{epsfig}
\usepackage{graphicx}
\usepackage{amsmath}
\usepackage{amssymb}
\usepackage{enumitem}
\usepackage{caption}
\usepackage{multirow, array, booktabs}
\usepackage{tabularx}
\usepackage{authblk}
\usepackage[skip=6pt plus1pt, indent=12pt]{parskip}

\usepackage[pagebackref=true,breaklinks=true,letterpaper=true,colorlinks,bookmarks=false]{hyperref}

\iccvfinalcopy 

\ificcvfinal\fi

\begin{document}

\title{Enhancing Landmark Detection in Cluttered Real-World Scenarios with Vision Transformers}


\author[ ]{Mohammad Javad Rajabi}
\author[ ]{Morteza Mirzai}
\author[ ]{Ahmad Nickabadi}
\affil[ ]{Amirkabir University of Technology (AUT), SML lab}
\affil[ ]{{\tt\small \{rajabi2001,mirzai,nickabadi\}@aut.ac.ir}}

\maketitle
\ificcvfinal\thispagestyle{empty}\fi

\begin{abstract}
Visual place recognition tasks often encounter significant challenges in landmark detection due to the presence of irrelevant objects such as humans, cars, and trees, despite the remarkable progress achieved by previous models, especially in the context of transformers. To address this issue, we propose a novel method that effectively leverages the strengths of vision transformers. By employing a meticulous selection process, our approach identifies and isolates specific patches within the image that correspond to occluding objects. To evaluate the efficacy of our method, we created augmented datasets and conducted comprehensive testing. The results demonstrate the superior accuracy achieved by our proposed approach. This research contributes to the advancement of landmark detection in visual place recognition and shows the potential of leveraging vision transformers to overcome challenges posed by cluttered real-world scenarios.
\end{abstract}

\section{Introduction}

Visual place recognition (VPR) is one of the core tasks of many promising applications in the field of computer vision\cite{netvlad,dislocation,cao2013graph,sattler2015hyperpoints,schindler2007city,Torii_2015_CVPR} and robotics\cite{chen2011city}, such as autonomous driving\cite{chalmers2017learning,mcmanus2014shady,mur2015orb}, geo-localization\cite{laskar2017camera,lim2012real}, 3D reconstruction\cite{chen2018encoder} and virtual reality\cite{middelberg2014scalable}. VPR in large-scale environments is usually defined as an image retrieval problem\cite{lowry2015visual}. Given a query image, the algorithm has to determine whether it is taken from a place already seen and identify the corresponding images from a database. However, this task is not without its challenges, as various factors such as different viewpoints, weather conditions, and illumination changes can introduce significant variations in the appearance of the same scene. In addition to viewpoint and weather challenges, one of the challenges faced by visual place recognition systems is the presence of occluding objects such as people or cars within the scene. These objects can obstruct the view of the environment, resulting in partial or distorted visual information that hinders accurate recognition. When occlusions occur, the recognition algorithms must be capable of handling incomplete or corrupted data to identify the location effectively.

\begin{figure}[!t]
\begin{center}
\includegraphics[width=0.45\textwidth]{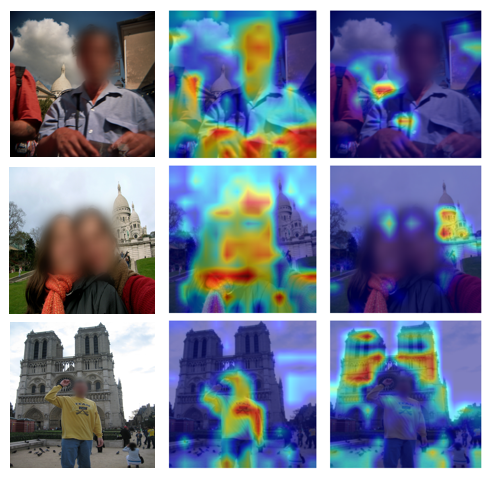}
\caption{Failure cases in VPR caused by occluding objects. The left column displays the original images, the middle column exhibits the saliency maps without the masking strategy, and the right column demonstrates the saliency maps with the implemented masking strategy. The presence of occluding objects hinders accurate landmark detection, as evident from the comparison between the middle and right columns.}\label{fig:fig1}
\vspace{-\baselineskip}
\end{center}
\end{figure}

Originally emerged in the field of natural language processing (NLP), the self-attention-based transformers have seen remarkable growth in the domain of computer vision as well. This trend began with the introduction of the vision transformer (ViT)\cite{dosovitskiy2020image}. Convolutional neural networks (CNNs) have historically been the dominant architecture for computer vision tasks, leveraging their ability to capture local patterns and spatial hierarchies effectively. Transformers, on the other hand, excel in modeling global dependencies and capturing long-range interactions through self-attention mechanisms. Previous studies \cite{song2023boosting,wang2022transvpr,zhu2023r2former} have shown the great potential of transformers in VPR, especially when the target landmark is clearly visible in the image. Nevertheless, in real-world scenarios, images are often cluttered with many irrelevant objects, such as humans, cars, and trees, which can interfere with landmark detection. Some models attempt to mitigate this problem by integrating local and global features \cite{cao2020unifying,tolias2020learning,yang2021dolg,radenovic2018fine} or employing attention mechanisms \cite{yuan2021softmp,song2022all,wang2022transvpr} to reduce the impact of occluding objects. However, in many cases, the occluding objects become exceedingly noticeable, overshadowing the intended landmark and resulting in the failure of the model. These distracting elements overpower the visual representation, thereby impeding the model's ability to accurately identify and handle the desired landmark.  Figure~\ref{fig:fig1} presents some failure cases caused by occluding objects

In this work, we focus on investigating the effectiveness of vision transformers in addressing the challenge of occluding objects in visual place recognition tasks. We propose a meticulous selection process, where we identify and isolate specific patches within the input image that correspond to occluding objects. To mitigate their impact, we employ sophisticated masking techniques to conceal these distracting elements from the model's attention. By implementing this strategy, we empower the model with the ability to accurately recognize the intended landmark, regardless of the size or influence of the occluding objects. Therefore, the targeted utilization of transformers in this manner yields significant improvements in performance and enhances the model's robustness when confronted with occlusions.

In order to evaluate the efficacy of the proposed approach, we have made augmented datasets by extending the well-known Paris6K\cite{paris} and Oxford5K\cite{oxford} landmark datasets. These augmented datasets pose notable challenges due to the inclusion of human subjects within each image, thereby introducing complexities that hinder accurate landmark recognition by models. The diverse range of poses and appearances exhibited by individuals within the augmented datasets introduces additional intricacies as well as a realistic benchmark which provides an opportunity to thoroughly examine the performance and robustness of models in challenging real-world scenarios

Our contributions can be summarized as follows:
\begin{enumerate}[noitemsep,nolistsep]
  \item Addressing the issue of occlusions in visual recognition tasks.
  \item Enhancing performance and robustness of transformers in the presence of occlusions by selectively masking occluding patches.
  \item Introducing augmented landmark datasets incorporating human subjects with diverse poses and appearances.
  \item Analysing the challenges posed by the augmented datasets and assessment of the model's performance on them.
\end{enumerate}

The rest of the paper is organized as follows. Section II reviews recent VPR studies. Section III describes the proposed model and Section IV gives the details of the generated augmented datasets. The results of the conducted experiments are reported in Section V. Finally, Section VI concludes this paper and gives suggestions for future works.

\section{Related Work}

In this section, we review previous works on place recognition, especially those related to transformers.
 
\textbf{Visual Place Recognition.} Traditionally, the large-scale VPR is framed as an image retrieval task, where the key is to find a discriminative representation for accurate and fast indexing. Several studies in image retrieval\cite{gordo2016deep,kalantidis2016cross,glv2,babenko2015aggregating,radenovic2018fine} are based on global descriptors. Common pooling methods are SPoC\cite{babenko2015aggregating}, CroW\cite{kalantidis2016cross}, R-MAC\cite{gordo2016deep}, and GeM\cite{radenovic2018fine}. More recently, attention-based methods are studied\cite{ng2020solar,song2022all}. Alternative approaches are to use local descriptors to re-rank according to geometry\cite{noh2017large,cao2020unifying,simeoni2019local} or build an aggregated representation\cite{radenovic2018revisiting,tolias2020learning}. The latter approaches have better performance but require more computational resources. It is also common to have interaction between local and global feature extraction\cite{cao2020unifying,yang2021dolg}.

\textbf{Attentions for Place Recognition.} In order to adaptively identify task-relevant regions in a complex scene image, an attention mechanism has been adopted in several VPR approaches. Among them, the learned attention maps can be considered as patch descriptor filters\cite{noh2017large,yuan2021softmp} or weight maps that modulate the CNN feature maps to generate global features\cite{chen2018learning,jin2017learned}. The attention module in CNN-based methods has usually been implemented as a shallow CNN which is trained separately\cite{noh2017large} or jointly\cite{cao2020unifying,chen2018learning,yuan2021softmp} with the backbone network.

\begin{figure*}[!tbp]
\setlength{\belowcaptionskip}{-20pt}
\begin{center}
    \includegraphics[width=0.75\linewidth]{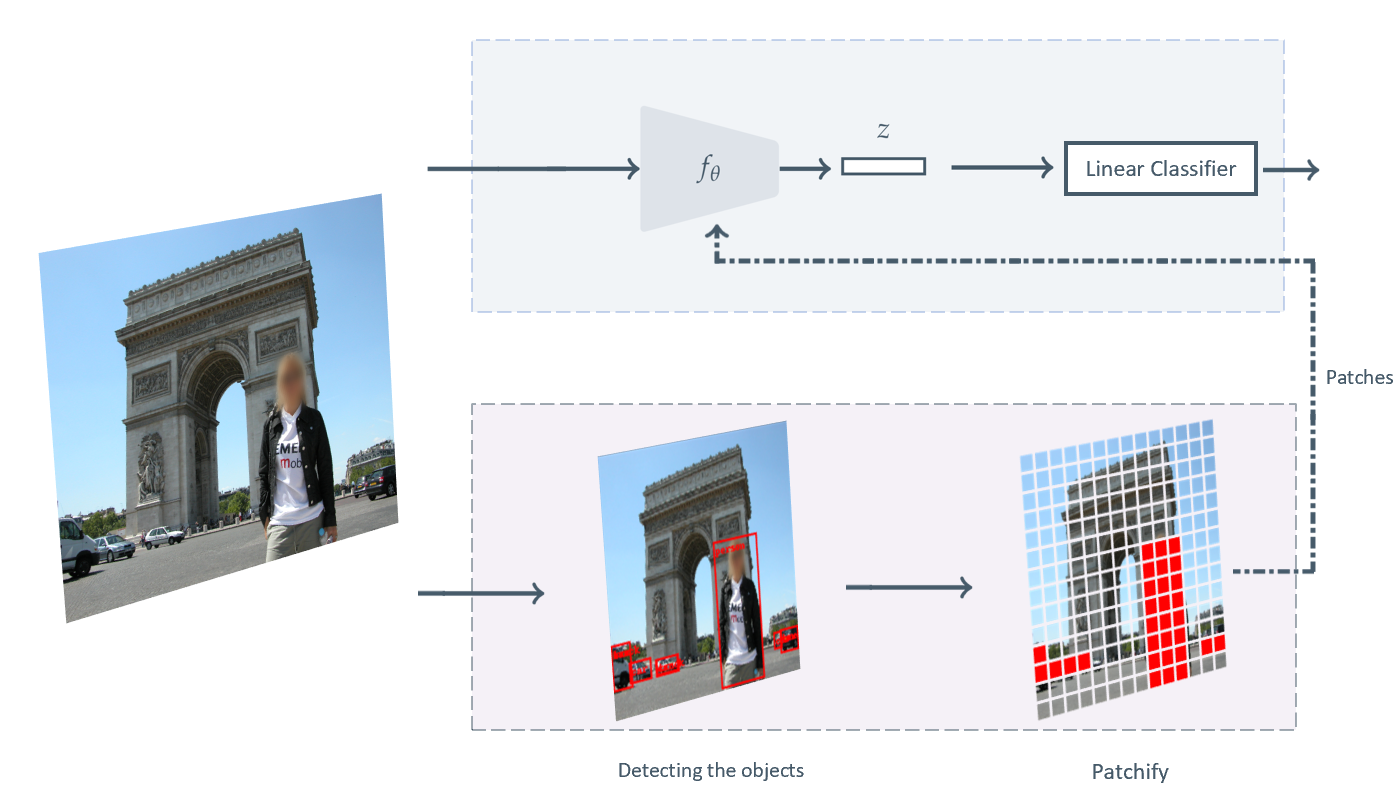}
  \caption{The overall architecture of the proposed method. Initially, the input image is passed through an object detector which generates bounding boxes encompassing occluding objects. These bounding boxes define the regions occupied by occluding patches within the image. Subsequently, the image, excluding the occluding patches, is fed into a transformer-based feature extractor, which extracts high-level features. By ignoring the occluding patches, the feature extractor focuses on the relevant information for landmark detection. Finally, a classifier employs the output feature vector to predict the labels corresponding to the landmarks. This comprehensive architecture ensures effective handling of occluding objects and facilitates accurate landmark recognition.\newline}
  \label{fig:arch}
\end{center}
\end{figure*}

\textbf{Vision Transformer.} Transformer architecture \cite{NIPS2017_3f5ee243} was first proposed for NLP tasks as a generic architecture without inductive bias. It is recently introduced to vision tasks as vision transformer\cite{dosovitskiy2020image} by simply considering each image patch as a token. The vanilla ViT\cite{dosovitskiy2020image} requires large-scale training datasets (e.g. ImageNet-1k\cite{deng2009imagenet}) to achieve comparable results as CNN\cite{he2016deep}. Deit\cite{touvron2021training} proposes a data-efficient training strategy for ViT which outperforms CNN on standard ImageNet-1k. The recent VG benchmark\cite{berton2022deep} adopts vanilla ViT without modifying the input resolution which could be suboptimal, while it already shows competitive performance on global retrieval.

\textbf{Transformer-based image retrieval.} Most of the transformer-based studies in computer vision use models pre-trained on large-scale datasets and apply them to downstream tasks such as object detection \cite{simeoni2021localizing,wang2023cut} and segmentation\cite{xie2021segformer,jain2023oneformer,zhang2022topformer}. There are only a few studies employing transformers for image retrieval. After initial off-the-shelf experiments\cite{gkelios2021investigating}, the image retrieval transformer (IRT)\cite{el2021training} has fine-tuned the model specifically for image retrieval\cite{radenovic2018fine}. The self-supervised regime is examined in DINO\cite{caron2021emerging}. The re-ranking transformer (RRT)\cite{tan2021instance} uses a transformer to re-rank images by local features, while super-features\cite{weinzaepfel2022learning} aggregate local features by an iterative transformer-inspired cross-attention mechanism; but in both cases, the features are still obtained by a convolutional network. No study has achieved performance competitive with convolutional networks so far.

\textbf{Datasets.} There are standard datasets commonly used for the evaluation of image retrieval techniques. Oxford5k\cite{oxford} has 5062 building images captured in Oxford with 55 query images. Paris6k\cite{paris} is composed of 6412 images of landmarks in Paris and also has 55 query images. These two datasets are often augmented with 100K distractor images from Flickr100k dataset\cite{oxford}, which constructs Oxford105k and Paris106k datasets, respectively. On the other hand, Holidays dataset\cite{jegou2008hamming} provides 1,491 images including 500 query images, which are from personal holiday photos. All these three datasets are fairly small, especially having a very small number of query images, which makes it difficult to generalize the performance tested in these datasets. Although Pitts250k\cite{torii2013visual} is larger, it is specialized to visual places with repetitive patterns and may not be appropriate for the general image retrieval task.

\section{The Proposed Method}

The overview of the proposed model is illustrated in Figure~\ref{fig:arch}. For a given input image, our method starts by feeding it into an object detector to identify patches that contain occluding objects. Subsequently, the original image, along with the masked patches, is passed through a transformer feature extractor. Finally, a classifier is employed to predict the label corresponding to the landmark depicted in the input image.

\textbf{Finding non-relevant patches.} To effectively mask specific regions of an image that are irrelevant or need to be removed, our approach incorporates an object detector. This component plays a crucial role in detecting and identifying objects present within the image. By analyzing the input image, the object detector generates bounding boxes that precisely outline the spatial location of the detected objects. The bounding box information obtained from the object detector is subsequently fed into the patchify module whose primary function is to extract patches from the image that correspond to the identified objects. By isolating these patches, we can focus our attention on specific areas of interest while disregarding unwanted objects that might be present in the image. By leveraging the object detector and patchify module together, our approach enables the selective masking of parts of the image. This targeted masking process ensures that the model can concentrate on relevant information while effectively ignoring or removing irrelevant objects.

\textbf{Transformer-based feature extractor.} Our approach exploits the potential of a pre-trained Vision Transformer model, which serves as our primary feature extractor. The ViT model has undergone extensive training on the ImageNet-1K\cite{deng2009imagenet} dataset. Leveraging the pre-trained ViT model grants us access to its learned high-level representations, capturing valuable features from input images efficiently. Utilizing ViT as our backbone provides a significant advantage in terms of speed and effectiveness in feature extraction.

The transformer network's inherent potential to selectively process arbitrary patches makes it an invaluable asset in image processing tasks. This unique capability empowers the model to prioritize specific areas of interest while disregarding irrelevant regions, resulting in enhanced focus on relevant features. The Vision Transformer model, known for its patch-based processing approach, exemplifies this versatility by effectively identifying and highlighting crucial information within images. By leveraging patch masking, we can mitigate the adverse effects of undesired objects, such as people or cars, present in the image. This strategic masking enables us to achieve improved accuracy and performance in our landmark detection task by effectively filtering out distractions and allowing the model to concentrate on the essential elements. We demonstrate that the transformer's ability to choose arbitrary patches, combined with the use of patch masking, provides a robust and efficient solution for enhancing the accuracy and effectiveness of image processing tasks.

To adopt the ViT model for our landmark detection task, we then fine-tune a classifier on the output features obtained from the ViT backbone. This fine-tuning process ensures that the classifier is specifically optimized to accurately classify the extracted features relevant to our landmark detection objective. By combining the robust feature extraction capabilities of ViT with the fine-tuned classifier, we achieve optimal performance in accurately identifying and classifying landmarks within our target dataset.

\section{Augmented Data Generation}

\begin{figure}[!t]
\begin{center}
\includegraphics[width=0.45\textwidth]{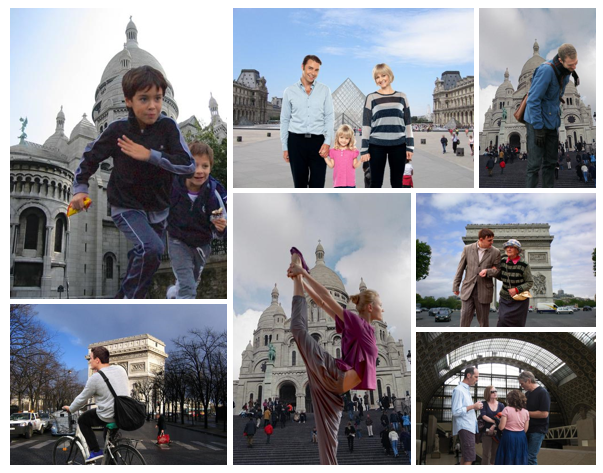}
\caption{Augmented1 dataset samples illustrating realistic poses and activities. The displayed samples highlight a wide range of poses and activities that closely resemble real-life scenarios.}\label{fig:aug1}
\vspace{-\baselineskip}
\end{center}
\end{figure}

\begin{figure}[!t]
\begin{center}
\includegraphics[width=0.45\textwidth]{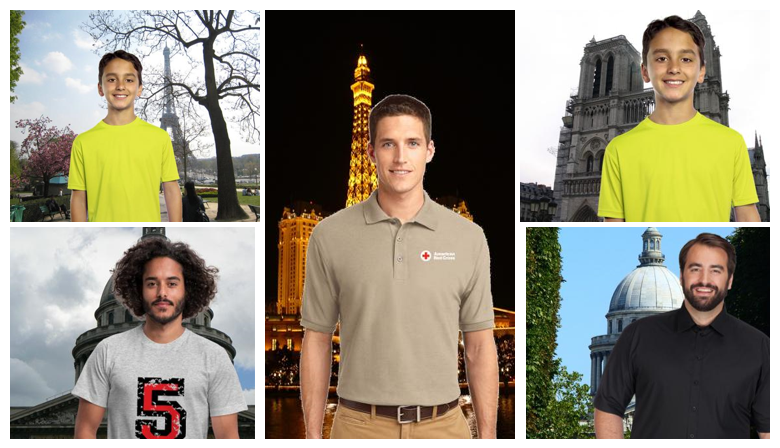}
\caption{Augmented2 dataset samples simulating real-life photography scenarios. These samples prominently feature individuals strategically positioned in front of the camera to recreate authentic real-life photography scenarios.}\label{fig:aug2}
\vspace{-\baselineskip}
\end{center}
\end{figure}

To assess the performance of our proposed landmark recognition model, we created two augmented datasets by building upon the existing Oxford5k\cite{oxford} and Paris6k\cite{paris} datasets. These augmented datasets were specifically assembled to evaluate the model's capability to accurately identify landmarks, even in the presence of distracting elements.
The first dataset, Augmented1, comprises individuals showcased in a range of poses and engaged in various activities, including running, sitting, and cycling with a bicycle. This augmentation introduces dynamic elements into the dataset, allowing us to assess the model's ability to recognize landmarks amidst varying human interactions. The second dataset, Augmented2, focuses on individuals positioned strategically in front of the camera, simulating the act of capturing a photograph with the featured building as the landmark. By including this synthetic dataset, we aim to evaluate the model's proficiency in accurately identifying landmarks, especially when confronted with compositions reminiscent of real-life photography scenarios. Example images of Augmented1 and Augmented2 datasets are presented in Figure~\ref{fig:aug1} and Figure~\ref{fig:aug2}, respectively.

By implementing our augmentation strategy, we substantially expanded the test dataset size by a factor of 20. This augmentation process involved selecting 20 distinct people images and skillfully overlaying them onto each test image in realistic positions. Consequently, we generated 20 new images for every original test image, effectively enhancing the diversity and variability within the dataset. This expanded dataset enables us to thoroughly evaluate the model's performance in handling a wide range of scenarios, capturing the intricacies of human presence in natural-looking compositions.

\section{Experiments}

The primary objective of our experiments revolves around assessing the efficacy of our proposed method in landmark recognition, employing a diverse range of datasets for comprehensive evaluation. Our main focus lies on the renowned Paris6k\cite{paris} and Oxford5k\cite{oxford} datasets, widely recognized in the field. However, to further challenge and validate the effectiveness of our method, we also incorporated two additional augmented datasets for each of the main datasets. This selection resulted in a total of six datasets, encompassing a rich variety of scenarios and complexities for a thorough evaluation.

To extract image features, we employed the vitl16 pre-trained MSN\cite{assran2022masked} model trained on ImageNet-1K\cite{deng2009imagenet}. Subsequently, we conducted training for a linear classifier over 50 epochs, utilizing a batch size of 256 images. Additionally, to accurately identify the patches in need of masking, we leveraged the Mask R-CNN\cite{he2017mask} model, a famous object detector famous for its precision. By utilizing this object detection model, we could effectively identify and isolate the regions of interest, enabling us to apply the necessary masking techniques. The combination of the vitl16 model for feature extraction and the Mask R-CNN\cite{he2017mask} model for patch identification formed a robust framework for our landmark recognition pipeline. The training process of the proposed scheme is carried out using a machine with an RTX 3000 GPU.

To ensure consistent feature extraction, we standardized the image size to 224 × 224 for the feature extractor, while the object detector operated on images of their original sizes. During the training phase, we focused on training the classifier using the main dataset without applying any masking techniques. This allowed us to establish a solid foundation for subsequent evaluations. However, during the evaluation process, we leveraged a masking procedure to eliminate the potential interference caused by humans and cars, thus enhancing the accuracy of landmark classification. Additionally, to thoroughly examine the robustness and versatility of our method, we introduced the challenging augmented datasets derived from the Paris6k\cite{paris} and Oxford5k\cite{oxford} datasets.

In order to identify potentially harmful patches within the images, we employed a two-step process. Initially, we utilized an object detector to scan the images and generate bounding boxes. These bounding boxes served as references for locating the patches of interest. However, it was observed that some patches did not completely fit within the bounding boxes. To address this, we employed a 70\% threshold to ensure that a bounding box was included as a masking patch, even if it only partially covered the patch. Once the patches requiring masking were identified, we combined them with the original image and fed the resulting data into the feature extractor. The feature representation obtained from the feature extractor was then utilized by the classifier to classify the image accurately. By incorporating this multi-step process, we aimed to effectively locate and mask the patches within the images that could potentially hinder accurate classification. This approach allowed us to leverage the strengths of both the object detector and the feature extractor, enabling our classifier to make informed decisions based on a comprehensive understanding of the images.

\subsection{Ablation Studies}

To show the effectiveness of our proposed method, we performed an ablation experiment. This experiment allowed us to analyze and quantify the contribution of different components in our approach. Specifically, we investigated the impact of the feature extractor and object detector on the performance and accuracy of the masking procedure. We also explored the effect of data augmentation techniques and the mask ratio. Our experiments yielded both quantitative and qualitative results that showcase the effectiveness and robustness of our method.

\textbf{Different pre-trained backbone:} ViT models come in different variants, such as small, base, and large, each with increasing complexity and parameterization as the model size grows. In our experiments, we explored all available pre-trained ViT models with a patch size of 16 to identify the most effective backbone for our method. Through careful analysis, we discovered interesting insights regarding the performance of different backbone models. In general scenarios, the base model (vitb) exhibited the best results. However, when it came to implementing the masking procedure, the large model (vitl) outperformed the others. This observation was reinforced by the conclusive findings presented in Table~\ref{tab:ablationl}, highlighting the superiority of the large model in handling the complexities introduced by the masking technique. 

\textbf{Different object detectors:} The object detector plays a pivotal role in our method, significantly influencing the overall accuracy and performance of our system. Understanding the impact of the object detector on our approach, we conducted a thorough evaluation by comparing the results obtained from two widely used object detection models: RetinaNet\cite{lin2017focal} and Mask R-CNN\cite{he2017mask}. These models were specifically assessed concerning their effectiveness in handling our two primary datasets, Paris6k\cite{paris} and Oxford5K\cite{oxford}. During our experiments, we discovered that the masking process itself had implications for performance. Particularly, we observed that certain patches, especially those located at the borders, exhibited poor performance when masked. This was attributed to the loss of crucial information necessary for accurate landmark recognition. The loss of this vital information impeded the system's ability to effectively identify and classify landmarks with precision.

\begin{table}[!t]
    \centering
    \resizebox{\linewidth}{!}{
    \begin{tabular}{c || c|c|c|c}
       \toprule[2.0pt]
       \multirow{2}{*}{Architecture} & \multicolumn{2}{c|}{Paris6K} & \multicolumn{2}{c}{Oxford5K} \\ \cline{2-5} 
       & w/ Masking & w/o Masking & w/ Masking & w/0 Masking \\ \midrule[0.2pt]
       Vits &  73.90 & 73.28 & 50.88 &  48.91   \\ 
       Vitb &  \textbf{75.31} & 74.53 & \textbf{52.81} & 52.86  \\ 
       Vitl (default)  & 75.15 &  \textbf{75.0} & 53.24 & \textbf{53.05}  \\  
       \bottomrule[2.0pt]
       \end{tabular}
     }
     \caption{Ablation study on ViT\cite{dosovitskiy2020image} with different backbone
      architectures on Paris6K\cite{paris} and Oxford5K\cite{oxford} datasets, with or without masking procedure.} 
     \label{tab:ablationl}
\end{table}

\begin{table}[!t]
    \centering
    \resizebox{\linewidth}{!}{
    \begin{tabular}{c || c|c}
       \toprule[2.0pt]
       Network  & Paris6K & Oxford5K \\ \midrule[0.2pt]  
       Vitl + RetinaNet &  73.43 &  49.7 \\ 
       Vitl +  Mask R-CNN (default)  & \textbf{74.37} & \textbf{52.26}  \\  
       \bottomrule[2.0pt]
        \end{tabular}
     }
     \caption{Ablation study on the impact of different object detectors on Paris6K\cite{paris} and Oxford5K\cite{oxford} datasets, with asking procedure.} 
     \label{tab:ablation2}
\end{table}

To address the challenges posed by masking and mitigate the potential loss of valuable information, we made a deliberate decision to utilize Mask R-CNN\cite{he2017mask} as our object detection model. As shown in Table~\ref{tab:ablation2}, this choice was driven by the unique capability of Mask R-CNN\cite{he2017mask} to provide us with precise object masks, which served as an additional source of information for selecting the appropriate patches. By leveraging both the bounding box and the exact mask, we selectively masked only the non-relevant patches. By focusing on masking only the non-relevant patches, we ensured that the essential visual cues necessary for accurate recognition of the landmark were retained.

\begin{table}[!t]
    \centering
    \resizebox{\linewidth}{!}{
    \begin{tabular}{c || c|c|c|c}
       \toprule[2.0pt]
       \multirow{2}{*}{Dataset} & \multicolumn{4}{c}{Mask Ratio} \\ \cline{2-5} 
       & 30 \% & 50 \% & 70 \% & 100 \% \\ \midrule[0.2pt]
       Paris6K &  74.06 & 74.06 & 74.37 &  74.68   \\ 
       Oxford5K &  51.47 & 51.47 & 52.26 &  52.07   \\ 
       \bottomrule[2.0pt]
       \end{tabular}
     }
     \caption{Ablation study on the impact of different values of mask ratio on Paris6K\cite{paris} and Oxford5K\cite{oxford} datasets.} 
     \label{tab:ablation3}
\end{table}

\begin{table*}[!t]
    \resizebox{\linewidth}{!}{
    \begin{tabular}{|c|c|c|c|c|c|c|}
    \hline
       \multirow{2}{*}{Network} & \multicolumn{3}{c|}{Paris6K} & \multicolumn{3}{c|}{Oxford5K} \\ \cline{2-7} 
       & Original & Augmented1 & Augmented2 & Original & Augmented1 & Augmented2 \\ \hline
       VGG16 & 66.88 & 47.11 & 42.21 & 43.98 & 27.80 & 26.15 \\ 
       ResNet50 &  70.31 & 25.85 & 25.82 & 47.14 & 25.08 &25.14   \\ 
       ViT &  \textbf{76.71} & 55.85 & 52.61 & \textbf{54.04} & 40.21 & 41.19\\ 
       MSN &  74.06 & 57.93 & 56.43 & 52.66 & 40.09 & 36.74\\ \hline \hline
       Ours &  74.37 & \textbf{62.5} & \textbf{64.23} & 52.26 & \textbf{42.78} & \textbf{42.06}\\ \hline 

        \end{tabular}
     }
     \caption{Comparison of our method with previous state-of-the-art results on the Original, Augmented1, Augmented2 of each Paris6K\cite{paris} and Oxford5K\cite{oxford} datasets. Our method leverages a large ViT\cite{dosovitskiy2020image} as the backbone of its feature extractor, combined with the proposed masking strategy.} 
     \label{tab:main}
\end{table*}

\textbf{Mask ratio.} In addition to the challenges we encountered, we faced the issue of objects within patches not completely filling the available space. If we were to choose a strategy of masking every patch that contained even a fraction of the object, we would risk losing a substantial amount of valuable information that is essential for the feature extractor. To overcome this challenge, we devised a solution by introducing a novel parameter known as the "mask ratio." This parameter played a vital role in determining whether a patch should be masked based on the percentage of object presence within it. We conducted a series of experiments to explore the impact of the mask ratio on the accuracy of our method. The results presented in Table~\ref{tab:ablation3} demonstrated that selecting the appropriate mask ratio had a significant influence on the overall performance. It became evident that the careful selection of the mask ratio was crucial for striking a balance between preserving the necessary object information and avoiding the loss of valuable details.

\subsection{Comparison with other models and methods}

To assess the efficacy of our proposed method, we conducted an extensive series of experiments, pitting it against other state-of-the-art models encompassing both convolutional neural network and vision transformer architectures. To ensure a fair and unbiased comparison, we meticulously adopted the same methodology for each model. This involved utilizing pre-trained models as feature extractors and training a linear classifier on our primary dataset.

During the training phase, all models showcased commendable performance, delivering promising results. However, the true differentiating factor emerged during the rigorous testing phase, where we subjected the models to our augmented datasets. It was during this critical evaluation that our method exhibited good promise, surging ahead while the CNN-based models, such as ResNet50\cite{he2016deep} and VGG16\cite{simonyan2014very}, experienced a substantial drop in accuracy. The results are shown in Table~\ref{tab:main}. Overall, the superior performance of our method on the augmented datasets solidifies its robustness and potential in challenging real-world scenarios.

Based on our analysis, it is evident that vision transformers exhibit strong performance on demanding landmark detection datasets. However, we have discovered that by employing our masking procedure, we can further capitalize on their inherent capabilities and effectively tackle common challenges encountered in landmark detection tasks, including the presence of humans and cars within the images. Our findings highlight the potential of the masking procedure to mitigate the impact of human and car interference.

It is worth noting that the pre-trained models available for vision transformers are primarily trained on general datasets like ImageNet-1k\cite{deng2009imagenet}. Unfortunately, there is a scarcity of pre-trained models specifically tailored for large landmark datasets such as Google-Landmarks\cite{glv2}. Consequently, in the interest of fairness and consistency, we evaluated and compared our method with models trained on the ImageNet-1k\cite{deng2009imagenet} dataset. Although these models may not be specifically optimized for landmark detection, our approach ensures a reliable test bed for the comparative evaluation. By utilizing the same dataset for pre-training and employing consistent evaluation metrics, we enable a fair assessment of the performance of our method against the existing models. While the availability of pre-trained models on landmark-specific datasets would be advantageous, the lack thereof necessitates our adherence to the ImageNet-1k\cite{deng2009imagenet} dataset for evaluation purposes. Despite this limitation, our evaluation and comparison remain valid within this context, allowing us to show the effectiveness of our method in relation to the available models.

\section{Conclusion}

This paper proposed a landmark recognition method as a robust and potent solution to the intricate challenges inherent in landmark detection. By seamlessly integrating a masking procedure, we have successfully harnessed the inherent strengths of vision transformers while surmounting common obstacles encountered in this complex task, particularly when faced with the presence of humans and cars within the images. We demonstrated the potential of our proposed masking strategy and vision transformers in overcoming complex visual contexts and providing valuable insights into the development of more refined and effective techniques for landmark detection. Furthermore, the success of our method can be attributed to its remarkable flexibility and versatility, which empowers it to surpass the boundaries of landmark detection and find applications in a diverse range of fields. 

{\small
\bibliographystyle{ieee_fullname}
\bibliography{egbib}
}

\end{document}